\documentclass[twocolumn,10pt]{IEEEtran}
\usepackage{amsmath,epsfig,epsfig,multirow,booktabs,amssymb,cite,mathtools,enumerate}

\usepackage{cite}

\begin{document}
\title{Ensemble Sparse Models for Image Analysis}

\author{Karthikeyan~Natesan~Ramamurthy, Jayaraman~J.~Thiagarajan, Prasanna~Sattigeri, and~Andreas~Spanias
\thanks{The authors are with the SenSIP Center, School of ECEE, Arizona State University, USA. 85287-5706}
\thanks{Email: \{knatesan, jjayaram, psattige, spanias\}@asu.edu}
}

% The paper headers
\markboth{IEEE Transactions on Image Processing}%
{}
% make the title area
\maketitle

\begin{abstract}
Sparse representations with learned dictionaries have been successful in several image analysis applications. In this paper, we propose and analyze the framework of ensemble sparse models, and demonstrate their utility in image restoration and unsupervised clustering. The proposed ensemble model approximates the data as a linear combination of approximations from multiple \textit{weak} sparse models. Theoretical analysis of the ensemble model reveals that even in the worst-case, the ensemble can perform better than any of its constituent individual models. The dictionaries corresponding to the individual sparse models are obtained using either random example selection or boosted approaches. Boosted approaches learn one dictionary per round such that the dictionary learned in a particular round is optimized for the training examples having high reconstruction error in the previous round. Results with compressed recovery show that the ensemble representations lead to a better performance compared to using a single dictionary obtained with the conventional alternating minimization approach. The proposed ensemble models are also used for single image superresolution, and we show that they perform comparably to the recent approaches. In unsupervised clustering, experiments show that the proposed model performs better than baseline approaches in several standard datasets.

\end{abstract}

\IEEEpeerreviewmaketitle

\section{INTRODUCTION}
\label{sec:intro}
Natural signals and images reveal statistics that allow them to be efficiently represented using a sparse linear combination of elementary patterns \cite{Field1987}. The local regions of natural images, referred to as patches, can be represented using a sparse linear combination of columns from a dictionary matrix. Given a data sample $\mathbf{x} \in \mathbb{R}^M$, and a dictionary matrix $\mathbf{D} \in \mathbb{R}^{M \times K}$, the data approximated using the linear generative model as $\mathbf{D} \mathbf{a}$, where  $\mathbf{a} \in \mathbb{R}^K$ is the sparse coefficient vector. This generative model that incorporates sparsity constraints in the coefficient vector, will be referred to as the \textit{sparse model}, in this paper. The dictionary can be either pre-defined or learned from the training examples themselves. Learning the dictionary will be alternatively referred to as learning the sparse model. Learned dictionaries have been shown to provide improved performance for restoring degraded data in applications such as denoising, inpainting, deblurring, superresolution, and compressive sensing \cite{Elad_book,JT_MLD}, and also in machine learning applications such as classification and clustering \cite{Wright,Ramirez,Mairal}. 

\subsection{Sparse Coding and Dictionary Learning}
\label{sec:SC_DL}
Using the linear generative model, the sparse code of a data sample $\mathbf{x}$ can be obtained by optimizing,
\begin{equation}
h(\mathbf{x},\mathbf{D}) = \min_{\mathbf{a}} \|\mathbf{x} - \mathbf{D}\mathbf{a}\|_2^2+\lambda \|\mathbf{a}\|_1.
\label{eqn:SC}
\end{equation} Here $ \|\mathbf{a}\|_1$ is the $\ell_1$ penalty that promotes sparsity of the coefficients, and the equivalence of (\ref{eqn:SC}) to $\ell_0$ minimization has been discussed in \cite{Donoho2003a} under some strong conditions on the dictionary $\mathbf{D}$. Some methods to obtain sparse representations used include the Matching Pursuit (MP) \cite{Mallat_MP}, Orthogonal Matching Pursuit (OMP) \cite{Tropp_Greed}, Order-Recursive Matching Pursuit \cite{cotter1999forward}, Basis Pursuit (BP) \cite{Chen_BP}, FOCUSS \cite{Gorodnitsky_Focuss} and iterated shrinkage algorithms \cite{Elad_2006_IS,Elad_2007_IS}. 

In several image processing andsample machine learning applications, it is advantageous to learn a dictionary, such that the set of training examples obtained from a probability space have a small approximation error with sparse coding. This problem can be expressed as minimizing the objective \cite{MairalOnline}
\begin{equation}
g(\mathbf{D}) = \mathbf{E}_\mathbf{x}[h(\mathbf{x},\mathbf{D})],
\label{eqn:DL_SC}
\end{equation} where the columns of $\mathbf{D}$, referred to as dictionary atoms, are  constrained to have unit $\ell_2$ norm, i.e., $\|\mathbf{d}_j\|_2 \leq 1, \forall j$. If the distribution in the probability space is unknown and we only have $T$ training examples $\{\mathbf{x}_i\}_{i=1}^T$, each with probability mass $p(\mathbf{x}_i)$, (\ref{eqn:DL_SC}) can be modified as the empirical cost function,
\begin{equation}
\hat{g}(\mathbf{D}) = \sum_{i=1}^T h(\mathbf{x}_i,\mathbf{D}) p(\mathbf{x}_i).
\label{eqn:DL_SC_emp}
\end{equation} 

Typically dictionary learning algorithms solve for the sparse codes \cite{LARS,FSS} using (\ref{eqn:SC}), and obtain the dictionary by minimizing $\hat{g}(\mathbf{D})$, repeating the steps until convergence. We refer to this baseline algorithm as \textit{Alt-Opt}. Since this is an alternating minimization process, it is important to provide a good initial dictionary and this is performed by setting the atoms to normalized cluster centers of the data \cite{KSVDElad}. Instead of learning dictionaries using sophisticated learning algorithms, it is possible to use the training examples themselves as the dictionary. Since the number of examples $T$ is usually much larger than the number of dictionary atoms $K$, it is much more computationally intensive to obtain sparse representations with examples. Nevertheless, both learned and example-based dictionaries have found applications in inverse problems \cite{Elad_denoise, Elad_book,yang2008image} and also in machine learning applications such as clustering and classification \cite{Aviyente2006, ThiagarajanAI, JT_radar,Wright,Ramirez,scspm,Yu,zhang,JT_LSC,JT_SLSC,Cheng2010}.

\subsection{Ensemble Sparse Models}
\label{sec:ensemble_sparse_model}
In this paper, we propose and explore the framework of ensemble sparse models, where we assume that data can be represented using a linear combination of $L$ different sparse approximations, instead of being represented using an approximation obtained from a single sparse model. The approximation to $\mathbf{x}$ can be obtained by optimizing
\begin{equation}
\min_{\{\beta_l\}_{l=1}^L}\|\mathbf{x} - \sum_{l=1}^L \beta_l \mathbf{D}_l \mathbf{a}_l\|_2^2.
\label{eqn:rep_ensemble}
\end{equation} Here each coefficient vector $\mathbf{a}_l$ is assumed to be sparse, and is obtained by solving for the optimization (\ref{eqn:SC}) with $\mathbf{D}_l$ as the dictionary. The weights $\{\beta_l\}_{l=1}^L$ control the contribution of each base model to the ensemble.

Since the ensemble combines the contributions of multiple models, it is sufficient that the dictionary for model is obtained using a  ``weak'' training procedure. We propose to learn these \textit{weak dictionaries} $\{\mathbf{D}_l\}_{l=1}^L$ sequentially, using a greedy forward selection procedure, such that training examples that incurred a high approximation error with the dictionary $\mathbf{D}_l$ are given more importance while learning $\mathbf{D}_{l+1}$. Furthermore, we also propose an ensemble model where each individual dictionary is designed as a random subset of training samples. The formulations described in this paper belong to the category of boosting \cite{Freund1999a} and random selection  algorithms \cite{polikar2006ensemble} in machine learning. In supervised learning, boosting is used to improve the accuracy of learning algorithms, using multiple weak hypotheses instead of a single strong hypothesis. The proposed ensemble sparse models are geared towards two image analysis problems, the inverse problem of restoring degraded images, and the problem of unsupervised clustering. Note that, boosted ensemble models have been used with the bag-of-words approach for updating codebooks in classification \cite{ZhangBoost} and medical image retrieval \cite{Wang_retrieval}. However, it has not been used so far in sparsity based image restoration problems or unsupervised clustering. Also when compared to \cite{elad2009plurality}, where the authors propose to obtain multiple randomized sparse representations from a single dictionary, in our approach, we propose to learn an ensemble of dictionaries and obtain a single representation from each of them. Typical ensemble methods for regression \cite{Duffy2002} modify the samples in each round of leveraging, whereas in our case the same training set is used for each round.

\subsection{Contributions}
\label{sec:contrib}
In this work, we propose the framework of ensemble sparse models and perform a theoretical analysis that relates their performance when compared to its constituent base sparse models. We show that, even in the worst case, an ensemble will perform at least as well as its best constituent sparse model. Experimental demonstrations that support this theory are also provided. We propose two approaches for learning the ensemble: (a) using a random selection and averaging (\textit{RandExAv}) approach, where each dictionary is chosen as a random subset of the training examples, and (b) using a boosted approach to learn dictionaries sequentially by modifying the probability masses of the training examples in each round of learning. In the boosted approach, two methods to learn the weak dictionaries for the individual sparse models, one that performs example selection using the probability distribution on the training set (\textit{BoostEx}), and the other that uses a weighted K-means approach (\textit{BoostKM}), are provided. For all cases of ensemble learning, we also provide methods to obtain the ensemble weights, $\{\beta_l\}_{l=1}^{L}$, from the training examples. Demonstrations that show the convergence of ensemble learning, with the increase in the number of constituent sparse models are provided. Experiments also show that the proposed ensemble approaches perform better than their best constituent sparse models, as predicted by theory.

In order to demonstrate the effectiveness of the proposed ensemble models, we explore its application to image recovery and clustering. The image recovery problems that we consider here are compressive sensing using random projections and single image superresolution. When boosted ensemble models are learned for image recovery problems, the form of degradation operator specific to the application is also considered, thereby optimizing the ensemble for the application. For compressive recovery, we compare the performance of the proposed Random Example Averaging (\textit{RandExAv}), Boosted Example (\textit{BoostEx}), and Boosted K-Means (\textit{BoostKM}) approaches to the single sparse model, whose dictionary is obtained using the \textit{Alt-Opt} approach. It is shown that the ensemble methods perform consistently better than a single sparse model at different number of measurements. Note that, the base sparse model for example-based approaches is designed as a random subset of examples, and hence it requires minimal training. Furthermore, in image superresolution, the performance of the proposed ensemble learning approaches is comparable to the recent sparse representation methods \cite{yang2010image}, \cite{yang2008image}.

Furthermore, we explore the use of the proposed approaches in unsupervised clustering. When the data are clustered along unions of subspaces, an $\ell_1$ graph \cite{Cheng2010} can be obtained by representing each data sample $\mathbf{x}_i$ as a sparse linear combination of the rest of the samples in the set. Another approach proposed in \cite{Ramirez} computes the sparse coding-based graph using codes obtained with a learned dictionary. We propose to use ensemble methods to compute sparse codes for each data sample, and perform spectral clustering using graphs obtained from them. Results with several standard datasets show that high clustering performance is obtained using the proposed approach when compared to $\ell_1$ graph-based clustering. 

\section{Analysis of Ensemble Models}
\label{sec:ensemble_analysis}
We will begin by motivating the need for an ensemble model in place of a single sparse model, and then proceed to derive some theoretical guarantees on the ensemble model. Some demonstrations on the performance of ensemble models will also be provided.

\subsection{Need for the Ensemble Model}
\label{sec:motivation}
In several scenarios, a single sparse model may be insufficient for representing the data, and using an ensemble model instead may result in a good performance. The need for ensemble models in supervised learning have been well-studied \cite{dietterich2000ensemble}. We will argue that the same set of reasons apply to the case of ensemble sparse models also. The first reason is statistical, whereby several sparse models may have a similar training error when learned from a limited number of training samples. However, the performance of each of these models with test data can be poor. By averaging representations obtained from an ensemble, we may obtain an approximation closer to the true test data. The second reason is computational, which can occur with the case of large training sets also. The inherent issue in this case is that sparse modeling is a problem with locally optimal solution. Therefore, we may never be able to reach the global optimal solution with a single model and hence using an ensemble model may result in a lesser representation error. Note that this case is quite common in dictionary learning, since many dictionary learning algorithms only seek a local optimal solution. The third reason for using an ensemble model is representational, wherein the hypothesis space assumed cannot represent the test data sample. In the case of sparse models, this corresponds to the case where the dictionary cannot provide a high-fidelity sparse approximation for a novel test data sample. This also happens in the case where the test observation is a corrupted version of the underlying test data, and there is ambiguity in obtaining a sparse representation. In this case also, it may be necessary to combine multiple sparse models to improve the estimate of the test data.

In order to simplify notation in the following analysis, let us denote the $l^{\text{th}}$ approximation in the ensemble model as  $\mathbf{c}_l = \mathbf{D}_l\mathbf{a}_l$.  The individual approximations are stacked in the matrix $\mathbf{C} \in \mathbb{R}^{M \times L}$, where $\mathbf{C} = [\mathbf{c}_1 \ldots \mathbf{c}_L]$ and the weight vector is denoted as $\boldsymbol{\beta} = [\beta_1 \ldots \beta_L]^T$. The individual residuals are denoted as $\mathbf{r}_l = \mathbf{x}-\mathbf{c}_l$, for $i = {1, \ldots, L}$, and the total residual of the approximation is given as
\begin{equation}
\label{eqn:res}
\mathbf{r} = \mathbf{x}-\mathbf{C}\boldsymbol{\beta}.
\end{equation} We characterize the behavior of the ensemble sparse model by considering four different cases for the weights $\{\beta_l\}_{l=1}^L$.

\subsubsection{Unconstrained Weights}
\label{sec:uncon_beta}
In this case, the ensemble weights $\{\beta_l\}_{l=1}^L$ are assumed to be unconstrained and computed using the unconstrained least squares approximation 
\begin{equation}
\label{eqn:uncon_ls}
\min_{\boldsymbol{\beta}} \|\mathbf{x}-\mathbf{C}\boldsymbol{\beta}\|_2^2
\end{equation}When the data $\mathbf{x}$ lies in the span of $\mathbf{C}$, the residual will be zero, i.e., $\mathbf{r} = 0$. The residual that has minimum energy in the $L$ approximations is denoted as $\mathbf{r}_{min}$. This residual can be obtained by setting the corresponding weight in the vector $\boldsymbol{\beta}$ to be $1$, whereas (\ref{eqn:uncon_ls}) computes $\boldsymbol{\beta}$ that achieves the best possible residual $\mathbf{r}$ for the total approximation. Clearly this implies
\begin{equation}
\label{eqn:res_bound}
\|\mathbf{r}\|_2 \leq \|\mathbf{r}_{min}\|_2.
\end{equation} Therefore, at worst, the total approximation will be as good as the best individual approximation.

\subsubsection{$\beta_l \geq 0$ }
\label{sec:nonneg_beta}
The ensemble weights $\{\beta_l\}_{l=1}^L$ are assumed to be non-negative in this case. The least squares approximation  (\ref{eqn:rep_ensemble}), with the constraint $\boldsymbol{\beta} \geq 0$ will now result in a zero residual if the data $\mathbf{x}$ lies in the simplical cone generated by the columns of $\mathbf{C}$. The simplical cone is defined as the set $\{\mathbf{b} : \mathbf{b} = \sum_{l=1}^L \mathbf{c}_l \beta_l\}$.  Otherwise, the bound on the total residual given by (\ref{eqn:res_bound}) holds in this case, since $\mathbf{r}_{min}$ can be obtained by setting the appropriate weight in $\boldsymbol{\beta}$ to $1$ in (\ref{eqn:res}), and the rest to $0$ in this case also.

\subsubsection{$\sum_{l=1}^L \beta_l = 1$ }
\label{sec:sumbeta}
When the ensemble weights are constrained to sum to $1$, the total residual can be expressed as 
\begin{equation}
\mathbf{r} = \sum_{l=1}^L \beta_l \mathbf{r}_l.
\label{eqn:res_aff_comb}
\end{equation}This can be easily obtained by replacing $\mathbf{x}$ as $\sum_{l=1}^L \beta_l \mathbf{x}$ in (\ref{eqn:res}). Denoting the residual matrix $\mathbf{R} = [\mathbf{r}_1 \ldots \mathbf{r}_L]$, the optimization (\ref{eqn:rep_ensemble}) to compute the weights can also be posed as $\min_{\boldsymbol{\beta}}\|\mathbf{R}\boldsymbol{\beta}\|_2$. Incorporating the constraint $\sum_{l=1}^L \beta_l = 1$, it can be seen that the final approximation $\mathbf{C}\boldsymbol{\beta}$ lies in the affine hull generated by the columns of $\mathbf{C}$, and the final residual, $\mathbf{R}\boldsymbol{\beta}$, lies in the affine hull generated by the columns of $\mathbf{R}$. Clearly the final residual will be zero, only if the data $\mathbf{x}$ lies in the affine hull of $\mathbf{C}$, or equivalently the zero vector lies in the affine hull of $\mathbf{R}$. When this does not hold, the worst case bound on $\mathbf{r}$ given by (\ref{eqn:res_bound}) holds in this case as well.

\subsubsection{$\beta_l \geq 0, \sum_{l=1}^L\beta_l = 1$ }
\label{sec:nonneg_sum_beta}
Similar to the previous case, the total residual can be expressed as (\ref{eqn:res_aff_comb}). As a result, the final representation $\mathbf{C}\boldsymbol{\beta}$ lies in the convex hull generated by the columns of $\mathbf{C}$, and the final residual, $\mathbf{R}\boldsymbol{\beta}$, lies in the convex hull generated by the columns of $\mathbf{R}$. Furthermore, the final residual will be zero only if the zero vector lies in the convex hull of $\mathbf{R}$. Clearly, the worst case bound on $\mathbf{r}$ given by (\ref{eqn:res_bound}) holds in this case. 

Although the worst case bounds for all the four cases are the same, the constraint spaces for the cases might provide us an idea about their relative performances with real data. The first case is unconstrained and it should result in the least error. The second case constrains that the solution should lie in the simplical cone spanned by the columns of $\mathbf{C}$, and this should lead to higher residual energy than Case 1. Case 3 constrains the solution to lie in an affine hull, which is of $L-1$ dimensions compared to simplical cone in $L$ dimensions, so it could lead to a higher error compared to Case 2. Case 4 is the subset of constraint spaces for Cases 1 to 3 and hence it will lead to the highest residual error. 

\begin{figure}[t]
  \centering
  \includegraphics[height=5cm]{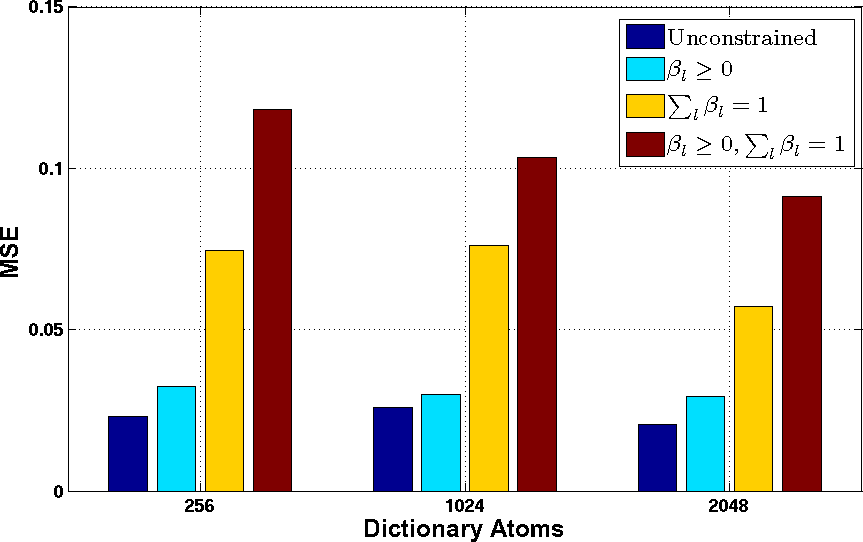}
\caption{Performance of the ``oracle'' ensemble models for various constraints on weights and different dictionary sizes in the base models.}
\label{Fig:four_oracles}
\end{figure}

\subsection{Demonstration of Ensemble Representations}
\label{sec:demo_ensemble}
In order to demonstrate the performance of ensemble representations with real data, we obtain a random set of $100,000$ patches, each of size $8 \times 8$, from a set of natural images. The training images were obtained from the superresolution toolbox published by Yang \textit{et. al.} \cite{YangSCSR_Code}, and consist of a wide variety of patterns and textures. We will refer to this set of training images simply as the \textit{training image set} throughout this paper. The chosen patches are then processed to remove the mean, followed by the removal of low-variance patches. Since image recovery is the important application of the proposed models, considering high-variance patches alone is beneficial. Each dictionary in the ensemble $\mathbf{D}_i$ is obtained as a random set of $K$ vectorized, and normalized patches. We fix the number of models in the ensemble as $L=20$. The test data is a random set of $1000$ grayscale patches obtained from the Berkeley segmentation dataset \cite{BSDS_dataset}. For each test sample, we compute the set of $L$ approximations using the sparse model given in (\ref{eqn:SC}), with $\lambda = 0.2$. The individual approximations are combined into an ensemble, under the four conditions on the weights, $\{\beta_l\}$, described above. The optimal weights are computed and the mean squared norm of the residuals for all the test samples are compared in Figure \ref{Fig:four_oracles}, for the dictionary sizes $K = \{256, 1025, 2048\}$. We observe that the performance of the ensembles generally improve as the size of the dictionaries used in the base models increase. The variation in  performance across all the four cases of weights follows our reasoning in the previous section. We refer to these as ``oracle'' ensemble models, since the weights are optimally computed with perfect knowledge of all the individual approximations and the actual data. In reality, the weights will be precomputed from the training data.

\section{PROPOSED ENSEMBLE SPARSE REPRESENTATION ALGORITHMS}
\label{sec:prop_algo}
The ensemble model proposed in (\ref{eqn:rep_ensemble}) results in a good approximation for any known data. However, in order to use ensemble models in analysis and recovery of images, that are possibly corrupted or degraded, both the weights $\{\beta_l\}_{l=1}^L$ and the dictionaries $\{\mathbf{D}_l\}_{l=1}^L$ must be inferred from uncorrupted training data. The set of weights is fixed to be common for all test observations instead of computing a new set of weights for each observation. Let us denote the set of training samples as $\mathbf{X} = [\mathbf{x}_1 \text{ } \mathbf{x}_2 \text{ } \ldots \text{ } \mathbf{x}_T]$,  and the set of coefficients in base model $l$ as $\mathbf{A}_l = [\mathbf{a}_{l,1} \text{ } \mathbf{a}_{l,2} \text{ } \ldots \text{ } \mathbf{a}_{l,T}]$, where $\mathbf{a}_{l,i}$ is the coefficient vector of the $i^{\text{th}}$ sample for base model $l$. In the proposed ensemble learning procedures, we consider both simple averaging and boosting approaches.

\subsection{Random Example Averaging Approach}
\label{sec:randex}
The first approach chooses $L$ random subsets of $K$ samples from the training data itself and normalizes them to form the dictionaries $\{\mathbf{D}_l\}_{l=1}^L$. The weights $\{\beta_l\}$ are chosen to be equal for all base models as $1/L$. Note that the selection of dictionaries follows the same procedure as given in the previous demonstration (Section \ref{sec:demo_ensemble}). We refer to this ensemble approach as \textit{Random Example Averaging} (\textit{RandExAv}).

\subsection{Boosting Approaches}
\label{sec:boost_ensemble}
The next two approaches use boosting and obtain the dictionaries and weights sequentially, such that the training examples that resulted in a poor performance with the ${l-1}^{\text{th}}$ base model are given more importance when learning the ${l}^{\text{th}}$ base model. We use a greedy forward selection procedure for obtaining the dictionaries and the weights. In each round $l$, the model is augmented with one dictionary $\mathbf{D}_l$, and the weight $\alpha_l$ corresponding to the dictionary is obtained. The cumulative representation for round $l$ is given by
\begin{equation}
\mathbf{X}_{l}  = (1-\alpha_l)\mathbf{X}_{l-1}+\alpha_l \mathbf{D}_l \mathbf{A}_l.
\label{eqn:greedy_forward}
\end{equation} Note that the weights of the greedy forward selection algorithm, $\alpha_l$, and the weights of the ensemble model, $\beta_l$, are related as
\begin{equation}
\beta_l = \alpha_l \prod_{t=l+1}^L (1-\alpha_t).
\label{eqn:alpha_beta}
\end{equation} From (\ref{eqn:greedy_forward}), it can be seen that $\mathbf{X}_{l}$ lies in the affine hull of $\mathbf{X}_{l-1}$ and $\mathbf{D}_l \mathbf{A}_l$. Furthermore, from the relationship between the weights $\{\alpha_l\}$
and $\{\beta_l\}$ given in (\ref{eqn:alpha_beta}), it is clear that $\sum_{l=1}^L \beta_l = 1$ and hence the ensemble model uses the constraints given in Case 3. Only the Cases 3 and 4 lead to an efficient greedy forward selection approach for the ensemble model in (\ref{eqn:rep_ensemble}), and we use Case 3 since it leads to a better approximation performance (Figure \ref{Fig:four_oracles}). 

In boosted ensemble learning, the importance of the training samples in a particular round is controlled by modifying their probability masses. Each round consists of (a) learning a dictionary $\mathbf{D}_{l}$ corresponding to the round, (b) computing the approximation for the current round $l$, (c) estimating the weight $\alpha_l$, (d) computing the residual energy for the training samples, and (e) updating the probability masses of the training samples for the next round. Since the goal of ensemble approaches is to have only \textit{weak} individual models, $\mathbf{D}_{l}$ is obtained using naive dictionary learning procedures as described later in this section. The dictionaries for the first round are obtained by fixing uniform probability masses for each training example in the first round, (i.e.),  $p_1(\mathbf{x}_i) = 1/T$ for $i = \{1,2,\ldots,T\}$. Assuming that $\mathbf{D}_{l}$ is known, the approximation for the current round is computed by coding the training samples $\mathbf{X}$ with the dictionary using (\ref{eqn:SC}). The weight $\alpha_l$ is computed such that the error between the training samples and the cumulative approximation $\mathbf{X}_{l}$ is minimized. Using (\ref{eqn:greedy_forward}), this optimization can be expressed as
\begin{equation}
\min_{\alpha_{l}} \left\|\mathbf{X}_{l} - [(1-\alpha_l)\mathbf{X}_{l-1}+\alpha_l \mathbf{D}_l \mathbf{A}_l]\right\|_F^2,
\label{eqn:comp_alpha}
\end{equation} and can be solved in closed form with the optimal value given as,
\begin{equation}
\alpha_{l} = \frac{\text{Tr}\left[(\mathbf{X}-\mathbf{X}_{l-1})^T(\mathbf{D}_l \mathbf{A}_l-\mathbf{X}_{l-1})\right]}{\left\|\mathbf{D}_l \mathbf{A}_l-\mathbf{X}_{l-1}\right\|_F^2},
\label{eqn:opt_alpha}
\end{equation} where \text{Tr} denotes the trace of the matrix. The residual matrix for all the training samples in round $l$ is given by $\mathbf{R}_l = \mathbf{X}-\mathbf{D}_l\mathbf{A}_l$. The energy of the residual for the $i^{\text{th}}$ training sample is given as $e_l(i) = \|\mathbf{r}_{l,i}\|_2^2$. If the dictionary in round $l$ provides a large approximation error for sample $i$, then that sample will be given more importance in round $l+1$. This will ensure that the residual error for sample $i$ in round $l+1$ will be small. The simple scheme of updating the probability masses as $p_{l+1}(\mathbf{x}_i) = e_l(i)$, upweights the badly represented samples and downweights the well-represented ones for the next round. 

Given a training set $\{\mathbf{x}_i\}_{i=1}^L$, and its probability masses $\{p_{l}(\mathbf{x_i})\}_{i=1}^L$, we will propose two simple approaches for learning the dictionaries corresponding to the individual sparse models.

\begin{figure}[t]
  \centering
  \includegraphics[height=5cm]{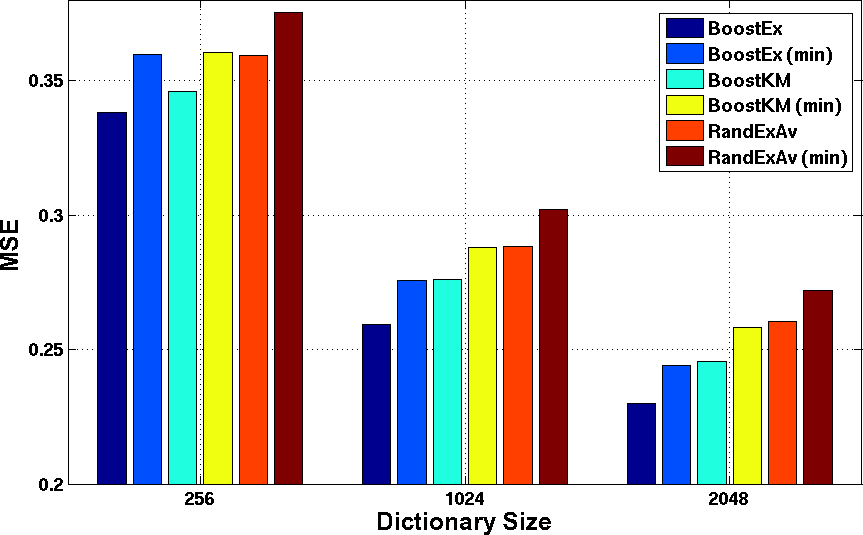}
\caption{Comparison of the performance of the proposed ensemble learning approaches for various dictionary sizes.}
\label{Fig:ensemble_dictsize}
\end{figure}

\subsubsection{BoostKM}
\label{sec:weighted_km}
When the sparse code for each training example is constrained to take one only one non-zero coefficient of value $1$, and the norms of the dictionary atoms are unconstrained, the dictionary learning problem (\ref{eqn:DL_SC_emp}) can be shown to reduce to K-Means clustering. Hence, computing a set of K-Means cluster centers and normalizing them to unit $\ell_2$ norm constitutes a reasonable weak dictionary. However, since the distribution on the data could be non-uniform in our case, we need to alter the clustering scheme to incorporate this. Denoting the cluster centers to be $\{\boldsymbol{\mu}_k\}_{k=1}^K$, the cluster membership sets to be $\{\mathcal{M}_k\}_{k=1}^K$, the weighted K-Means objective is denoted as
\begin{equation}
\min_{\{\boldsymbol{\mu}_k\}_{k=1}^K, \{\mathcal{M}_k\}_{k=1}^K} \sum_{k=1}^K \sum_{i \in \mathcal{M}_k} p(\mathbf{x}_i) \|\mathbf{x}_i - \boldsymbol{\mu}_k\|_2^2.
\label{eqn:weight_km}
\end{equation}The weighted K-Means procedure is implemented by modifying the scalable K-Means++ algorithm, also referred to as the K-Means$\parallel$ (K-Means Parallel) algorithm \cite{bahmani2012scalable}. The K-Means$\parallel$ algorithm is an improvement over the K-Means++ algorithm \cite{arthur2007} that provides a method for careful initialization leading to improved speed and accuracy in clustering. The advantage with the K-Means$\parallel$ algorithm is that the initialization procedure is scalable to a large number of samples. In fact, it has been shown in \cite{bahmani2012scalable} that just the initialization procedure in K-Means$\parallel$ results in a significant reduction in the clustering cost. Since we are interested in learning only a weak dictionary, we will use the normalized cluster centers obtained after initialization as our dictionary. The K-Means++ algorithm selects initial cluster centers sequentially such that they are relatively spread out. For initializing $K$ cluster centers, the algorithm creates a distribution on the data samples and picks a cluster center by sampling it and appends it to the current set of centers. The distribution is updated after each cluster center is selected. In contrast, the K-Means$\parallel$ algorithm updates the distribution much more infrequently, after choosing $q$ cluster centers in each iteration. This process is repeated for $s$ iterations, and finally the number of cluster centers obtained is $sq$. The chosen centers are re-clustered to obtain the initial set of $K$ clusters. It is clear that $s$ must be chosen such that $sq>K$. We provide only the initialization of the weighted K-Means$\parallel$ algorithm that takes the data distribution, $\{p_l(\mathbf{x}_i)\}_{i=1}^T$, also into consideration.

Let us denote $\delta_i$ as the shortest distance of the $i^{\text{th}}$ training sample to the set of cluster centers already chosen. The initialization of the weighted K-Means$\parallel$ algorithm proceeds as follows:
\begin{enumerate}[(a)]
\item{Initialize $\overline{\mathcal{M}} = \{ \}$.}
\item{Pick the first center $\boldsymbol{\mu}_1$ from the training set based on the distribution $\{p_l(\mathbf{x}_i)\}_{i=1}^T$, and append it to $\overline{\mathcal{M}}$.}
\item{The set of intermediate cluster centers, $\mathcal{M}'$, is created using $q$ samples from the data, $\{\mathbf{x}_i\}_{i=1}^T$, according to the probability $\frac{p_l(\mathbf{x}_i) \delta_i^2} {\sum_{j=1}^T p_l(\mathbf{x}_j) \delta_j^2}$.}
\item{Augment the set $\overline{\mathcal{M}} \leftarrow \overline{\mathcal{M}} \cup \mathcal{M}'$.}
\item{Repeat steps 2, 3 and 4 for $s$ iterations.}
\item{Set the weight of each element $\boldsymbol{\mu}$ in the set $\overline{\mathcal{M}}$, as the sum of weights of samples in $\mathbf{X}$ that are closer to $\boldsymbol{\mu}$ than any other sample in $\overline{\mathcal{M}}$.}
\item{Perform weighted clustering on the elements of $\overline{\mathcal{M}}$ to obtain the set of $K$ cluster centers, $\mathcal{M}$.}
\end{enumerate} Note that the steps (b) and (c) are used to compute the initial cluster centers giving preference to samples with higher probability mass. Finally, each dictionary atom $\mathbf{d}_k$ is set as the normalized cluster center $\frac{\boldsymbol{\mu}_k}{\|\boldsymbol{\mu}_k\|_2}$.

\subsubsection{BoostEx}
\label{sec:ex_dl}
From (\ref{eqn:DL_SC_emp}), it is clear that the learned dictionary atoms are close to training samples that have higher probabilities. Therefore, in the \textit{BoostEx} method, the dictionary for round $l$ is updated by choosing $K$ data samples based on the non-uniform weight distribution, $\{p_l(\mathbf{x}_i)\}_{i=1}^T$, and normalizing them. This scheme will ensure that those samples with high approximation errors in the previous round, will be better represented in the current round. 

\begin{figure}[t]
  \centering
  \includegraphics[height=5cm]{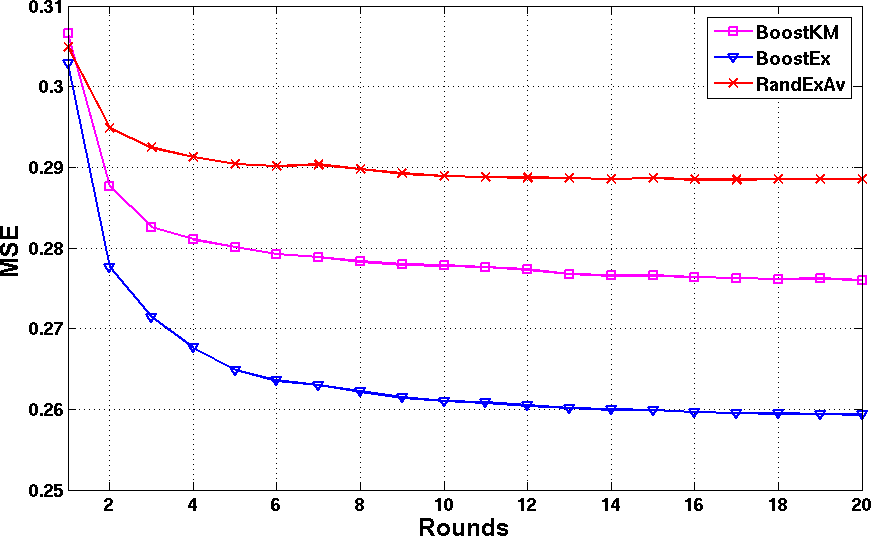}
\caption{Convergence characteristics of the proposed ensemble learning approaches.}
\label{Fig:ensemble_1024}
\end{figure}

\begin{figure*}
\begin{minipage}[b]{1\linewidth}
 \centering
 \includegraphics[width = 13cm]{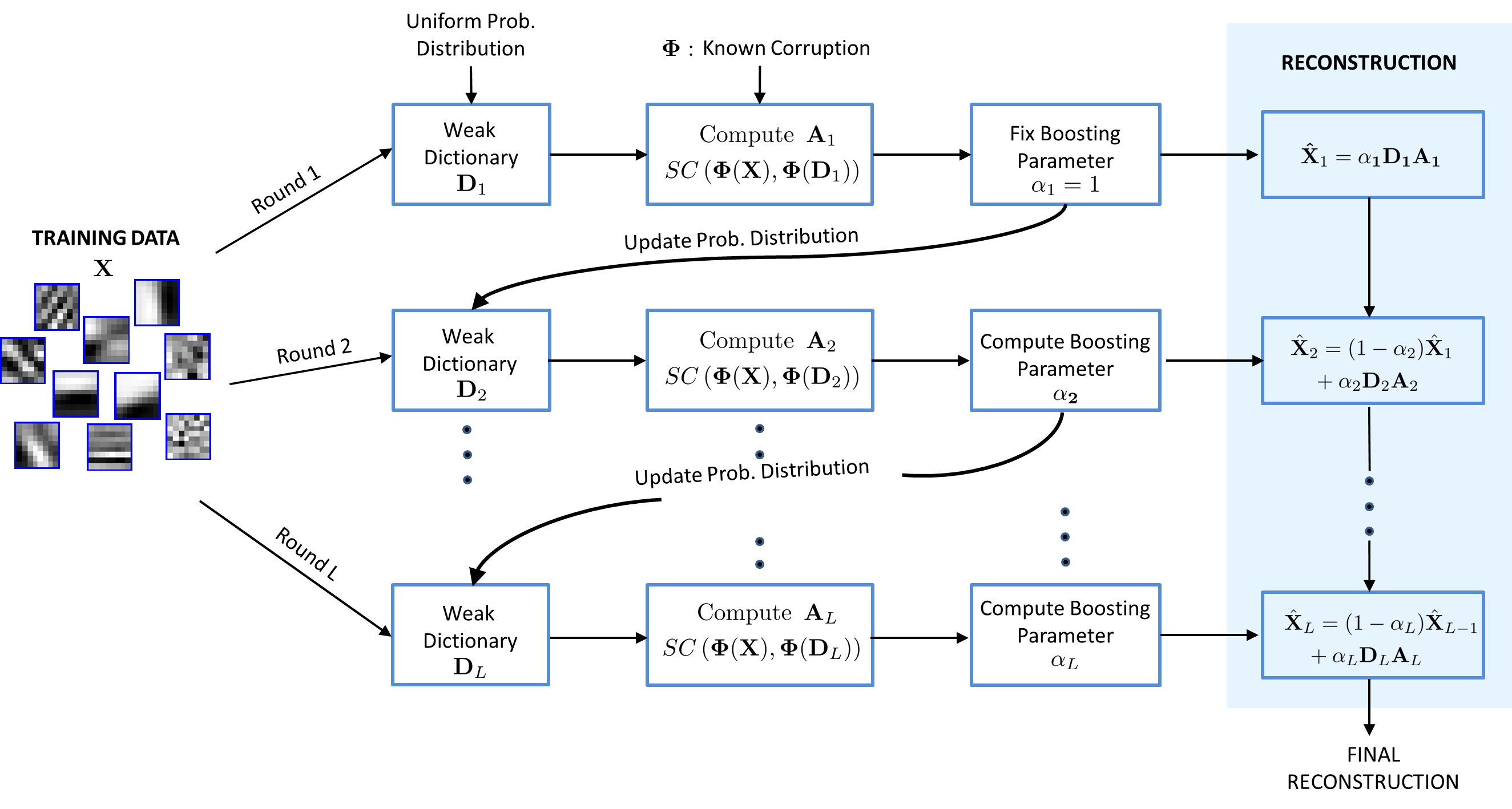}
% \vspace{1.5cm}
\end{minipage}
\caption{Illustration of the proposed boosted dictionary learning for image restoration. SC denotes sparse coding using (\ref{eqn:SC_degrade}).}
\label{Fig:BoostDL}
\end{figure*}

\subsection{Demonstration of the Proposed Approaches}
\label{sec:demo_prop_ensemble}
The performance of the proposed ensemble schemes for dictionaries of three different sizes $K = \{256,1024,2048\}$ are compared. The training set described in Section \ref{sec:demo_ensemble} is used with the \textit{RandExAv}, \textit{BoostKM}, and \textit{BoostEx} schemes. The dictionaries $\{\mathbf{D}_l\}_{l=1}^L$ and the weights $\{\beta_l\}_{l=1}^L$ are obtained with the above schemes for $L=20$. The individual approximations in the training set are obtained using (\ref{eqn:SC}) with the sparsity penalty set as $\lambda = 0.2$.  For each sample in the test set described in Section \ref{sec:demo_ensemble}, the individual representations are computed using (\ref{eqn:SC}) with $\lambda = 0.2$. The ensemble approximation the $i^{\text{th}}$ test sample is obtained as $\sum_{l=1}^L \beta_l \mathbf{D}_l\mathbf{a}_{l,i}$. Figure \ref{Fig:ensemble_dictsize} compares the performances of the proposed schemes for different dictionary sizes. The minimum error obtained across all individual approximations is also shown for comparison, with all the three methods and the different dictionary sizes. It can be seen that the proposed schemes satisfy the basic property of the ensemble discussed in Section \ref{sec:ensemble_analysis}, where it has been shown that the ensemble approximation performs better than the best constituent individual approximation. As the number of number of approximations in the ensemble increase, the average mean squared error (MSE) for the three proposed methods reduce, as shown in Figure \ref{Fig:ensemble_1024} for a dictionary size of $1024$. Clearly, increasing the number of models in the ensemble results in a better approximation, but the MSE flattens out as the number of rounds increase.

\section{Application: Image Restoration}
\label{sec:im_restore}
In restoration applications, it is necessary to solve an inverse problem, in order to estimate the test data $\mathbf{y}$ from
\begin{equation}
\mathbf{z} =  \mathbf{\Phi} (\mathbf{y})+\mathbf{n},
\label{eqn:degrade}
\end{equation} where $\mathbf{\Phi}(.)$ is the corruption operator and $\mathbf{n}$ is the additive noise. If the operator $\mathbf{\Phi}(.)$ is linear, we can represent it using the matrix $\mathbf{\Phi}$. With the prior knowledge that $\mathbf{y}$ is sparsely representable in a dictionary $\mathbf{D}$ according to (\ref{eqn:SC}), (\ref{eqn:degrade}) can be expressed as $\mathbf{z} =  \mathbf{\Phi}\mathbf{D}\mathbf{a}+\mathbf{n}$. Restoring $\mathbf{x}$ now reduces to computing the sparse codes $\mathbf{a}$ by solving 
\begin{equation}
\min_{\mathbf{a}} \|\mathbf{z} - \mathbf{\Phi}\mathbf{D}\mathbf{a}\|_2^2+\lambda \|\mathbf{a}\|_1.
\label{eqn:SC_degrade}
\end{equation} and finally estimating $\mathbf{y} = \mathbf{D} \mathbf{a}$ \cite{Elad_book}. In the proposed ensemble methods, the final estimate of $\mathbf{x}$ is obtained as a weighted average of the individual approximations. Furthermore, in the boosting approaches, \textit{BoostKM} and \textit{BoostEx}, the degradation operation can be included when learning the ensemble. This is achieved by degrading the training data as $\mathbf{\Phi}\mathbf{X}$, and obtaining the approximation with the coefficients computed using (\ref{eqn:SC_degrade}) instead of (\ref{eqn:SC}). The procedure to obtain boosted dictionaries using degraded data and computing the final approximation is illustrated in Figure \ref{Fig:BoostDL}. In this figure, the final approximation is estimated sequentially using the weights $\{\alpha_l\}_{l=1}^L$, but it is equivalent to computing $\{\beta_l\}_{l=1}^L$ using (\ref{eqn:alpha_beta}) and computing the ensemble estimate $\sum_{l=1}^L \beta_l \mathbf{D}_l \mathbf{A}_l$.

\begin{table*}[htbp]
\setlength{\tabcolsep}{4pt}
  \centering
  \caption{Compressed recovery of standard images: PSNR (dB) obtained using Alternating Dictionary Optimization (\textit{Alt-Opt}), BoostEx (\textit{BEx}), BoostKM (\textit{BKM}), RandExAv (\textit{RExAv}), and \textit{Ex-MLD} methods, for different values of $N$. The results reported were obtained by averaging over $10$ iterations with different random measurement matrices. In each, the  higher PSNR is given in bold font.}
      \footnotesize	
    \begin{tabular}{c|ccccc|ccccc|ccccc}
    \toprule
    \multicolumn{1}{c|}{\multirow{2}[0]{*}{\textbf{Image}}} & \multicolumn{15}{c}{\textbf{Number of Measurements (N)}} \\
    \multicolumn{1}{c|}{} & \multicolumn{5}{c}{\textbf{N = 8}} & \multicolumn{5}{c}{\textbf{N = 16}} & \multicolumn{5}{c}{\textbf{N = 32}} \\
    \midrule
          & \textbf{Alt-Opt} & \textbf{BEx} & \textbf{BKM} & \textbf{RExAv} & \textbf{Ex-MLD} & \textbf{Alt-Opt} & \textbf{BEx} & \textbf{BKM} & \textbf{RExAv} & \textbf{Ex-MLD} & \textbf{Alt-Opt} & \textbf{BEx} & \textbf{BKM} & \textbf{RExAv}&\textbf{Ex-MLD}  \\
    Barbara & 21.55 & 22.05 & 22.04 & 22.08&\textbf{22.95} & 23.52 & 23.86 & 23.73 & 23.68 & \textbf{24.39} & 26.45 & 26.53 & 26.44 & 26.28& \textbf{26.66} \\
    Boat  & 23.08 & 23.73 & 23.95 & 23.99&\textbf{25.08} & 25.91 & 26.29 & 26.57 &26.59& \textbf{26.96} & 28.79 & 29.32 & 29.61 & 29.61&\textbf{29.9} \\
    Couple & 23.15 & 23.81 & 24.02 & 24.05&\textbf{25} & 25.87 & 26.30 & 26.57 & 26.56&\textbf{27.19} & 28.83 & 29.33 & 29.68 & 29.67&\textbf{29.8} \\
    Fingerprint & 18.10 & 18.76 & 19.15 & 19.16& \textbf{20.39} & 21.74 & 22.18 & 22.84 &22.86& \textbf{23.19} & 25.36 & 25.85 & 26.59 & 26.63&\textbf{26.82} \\
    House & 24.52 & 25.12 & 25.51 & 25.52&\textbf{26.55} & 28.01 & 28.14 & 28.63 & 28.66&\textbf{28.93} & 31.28 & 31.53 & 32.01 & 32.03&\textbf{32.25} \\
    Lena  & 25.14 & 25.84 & 26.18 &26.25& \textbf{27.17} & 28.31 & 28.73 & 29.08 & 29.13&\textbf{29.59} & 31.12 & 31.77 & 32.15 &32.17& \textbf{32.65} \\
    Man   & 23.90 & 24.60 & 24.83 & 24.89&\textbf{25.84} & 26.60 & 27.12 & 27.35 & 27.40 & \textbf{27.68}& 29.45 & 30.14 & 30.40 & 30.42 & \textbf{30.67} \\
    Peppers & 21.31 & 21.83 & 22.17 & 22.23 & \textbf{23.12} & 24.30 & 24.54 & 24.82 & 24.91&\textbf{25.68} & 27.28 & 27.69 & 28.03 & 28.11&\textbf{28.57} \\
    \bottomrule
    \end{tabular}%
  \label{table:cs_perf}%
\end{table*}%

\normalsize

\begin{figure*}[t]
\begin{minipage}[c]{0.19\linewidth}
 \centering
 \includegraphics[width = 3.5cm]{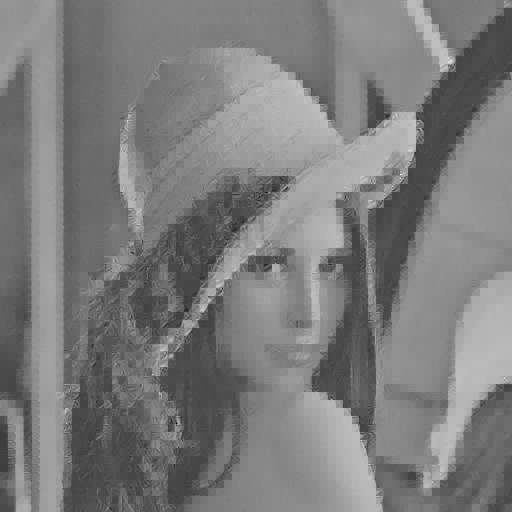}
 \centerline{{\textit{Alt-Opt} (24.81 dB)}}\medskip
\end{minipage}
\hfill\begin{minipage}[c]{0.19\linewidth}
 \centering
 \includegraphics[width = 3.5cm]{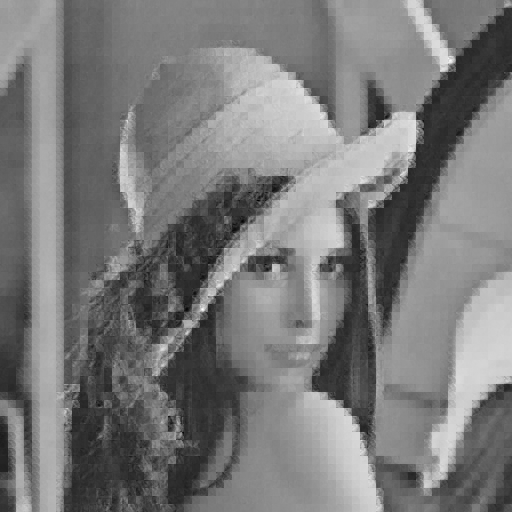}
 \centerline{{\textit{BEx} (25.7 dB)}}\medskip
\end{minipage}
\hfill\begin{minipage}[c]{0.19\linewidth}
 \centering
 \includegraphics[width = 3.5cm]{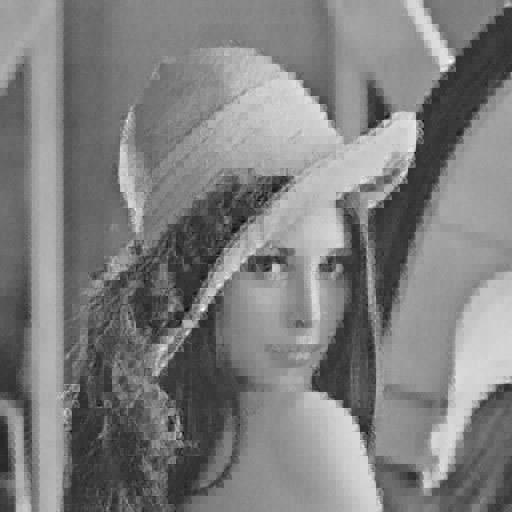}
 \centerline{{\textit{BKM} (26.06 dB)}}\medskip
\end{minipage}
\hfill\begin{minipage}[c]{0.19\linewidth}
 \centering
 \includegraphics[width = 3.5cm]{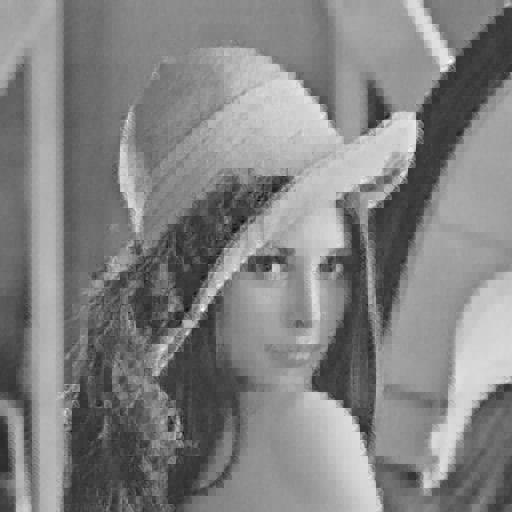}
 \centerline{{\textit{RExAv} (26.15 dB)}}\medskip
\end{minipage}
\hfill\begin{minipage}[c]{0.19\linewidth}
 \centering
 \includegraphics[width = 3.5cm]{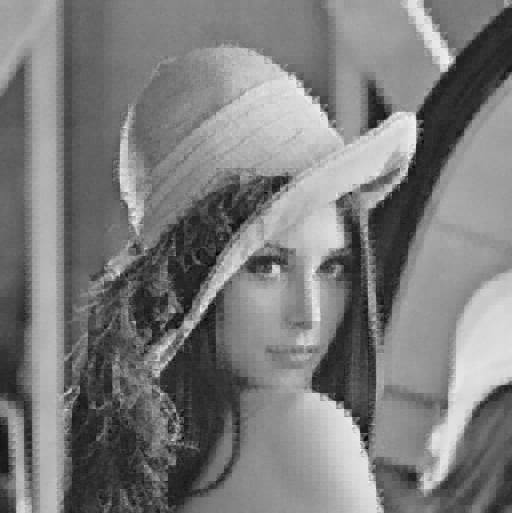}
 \centerline{{\textit{Ex-MLD} (27.19 dB)}}\medskip
\end{minipage}
\caption{Compressed recovery of \textit{Man} image using \textit{BoostKM} dictionaries. The reconstructed images along with their corresponding PSNR are shown for the rounds $\{1, 5, 20, 50\}$, when $25\%$ random measurements are used.}
\label{Fig:kmboost_cs_recim}
\end{figure*}

\subsection{Compressive Recovery}
In compressed sensing (CS), the $N-$dimensional observation $\mathbf{z}$ is obtained by projecting the $M-$dimensional data $\mathbf{y}$ onto a random linear subspace, where $N \ll M$ \cite{Donoho2006}. In this case, the entries of the degradation matrix $\mathbf{\Phi} \in \mathbb{R}^{N \times M}$ are obtained as i.i.d. realizations of a Gaussian or Bernoulli random variable. Compressive recovery can be effectively performed using conventional dictionaries or ensemble dictionaries. In addition, the proposed idea of ensemble learning can be incorporated in existing learning schemes to achieve improved recovery performance. In particular, the multilevel dictionary learning algorithm \cite{JT_MLD} can be very easily adapted to compute ensemble representations. Before discussing the experimental setup, and the results of the proposed methods, we will describe the modification to multilevel dictionary learning for improving the compressed recovery performance with learned dictionaries.

\subsubsection{Improved Multilevel Dictionaries} 
The multilevel dictionary (MLD) learning algorithm is a hierarchical procedure where the dictionary atoms in each level are obtained using a 1-D subspace clustering procedure \cite{JT_MLD}. Multilevel dictionaries have been shown to generalize well to novel test data, and have resulted in high performance in compressive recovery. We propose to employ the \textit{RandExAv} procedure in each level of multilevel learning to reduce overfitting and thereby improve the accuracy of the dictionaries in representing novel test samples. In each level, $L$ different dictionaries are drawn as random subsets of normalized training samples. For each training sample, a $1-$sparse representation is computed with each individual dictionary, and the approximations are averaged to obtain the ensemble representation for that level. Using the residual vectors as the training data, this process is repeated for multiple levels. The sparse approximation for a test sample is computed in a similar fashion. Since the sparse code computation in each level is performed using simple correlation operations, the computation complexity is not increased significantly by employing ensemble learning. In our simulations, we will refer to this approach as Example-based Multilevel Dictionary learning (\textit{Ex-MLD}).

\subsubsection{Results}
The training set is the same as that described in Section \ref{sec:demo_ensemble}. For the baseline \textit{Alt-Opt} approach, we train a single dictionary with $K = 256$ using $100$ iterations with the sparsity penalty $\lambda_{tr}$ set to $0.1$. The ensemble learning procedures \textit{BoostEx}, \textit{BoostKM} and \textit{RandExAv} are trained with $L = 50$ and $K = 256$ for sparsity penalty $\lambda_{tr} = 0.1$. The boosted ensembles are trained by taking the random projection operator into consideration, as discussed in Section \ref{sec:im_restore} for the reduced measurements, $N = \{8, 16, 32\}$. For the \textit{Ex-MLD} method, both the number of levels and the number of atoms in each level were fixed at $16$. In each level, we obtained $L = 50$ dictionaries to compute the ensemble representation.

The recovery performance of the proposed ensemble models is evaluated using the set of standard images shown in Table \ref{table:cs_perf}. Each image is divided into non-overlapping patches of size $8 \times 8$, and random projection is performed with the number of measurements set at $N = \{8, 16, 32\}$.  For the \textit{Alt-Opt} procedure, the individual patches are recovered using (\ref{eqn:SC_degrade}), and for the ensemble methods, the approximations computed using the $L$ individual dictionaries are combined. The penalty $\lambda_{te}$ is set to $0.1$ for sparse coding in all cases. For each method, the PSNR values were obtained by averaging the results over $10$ iterations with different random measurement matrices, and the results are reported in Table \ref{table:cs_perf}. It was observed that the proposed ensemble methods outperform the \textit{Alt-Opt} methods in all cases. In particular, we note that the simple \textit{RandExAv} performs better than the boosting approaches, although in Section \ref{sec:demo_ensemble} it was shown that boosting approaches show a superior performance. The reason for this discrepancy is that boosting aggressively reduces error with training data, and hence may lead to overfitting with degraded test data. Whereas, the \textit{RandExAv} method provides the same importance to all individual approximations both during the training and the testing phases. As a result, it provides a better generalization in the presence of degradation. We also note that similar behavior has been observed with ensemble classification methods \cite{dietterich2000experimental}. Random sampling methods such as bagging perform better than boosting with noisy examples, since bagging exploits classification noise to produce more diverse classifiers. Furthermore, we observed that the proposed \textit{Ex-MLD} method performed significantly better than all approaches, particularly for lower number of measurements. Figure \ref{Fig:kmboost_cs_recim} shows the images recovered using the different approaches, when $N$ was fixed at $8$. As it can be observed, the \textit{Ex-MLD} and \textit{RandExAv} methods provide PSNR gains of $2.38$dB and $1.34$dB respectively, when compared to the \textit{Alt-Opt} approach.  

\begin{figure}[t]
\begin{minipage}[c]{1.0\linewidth}
 \centering
 \includegraphics[width = 8cm]{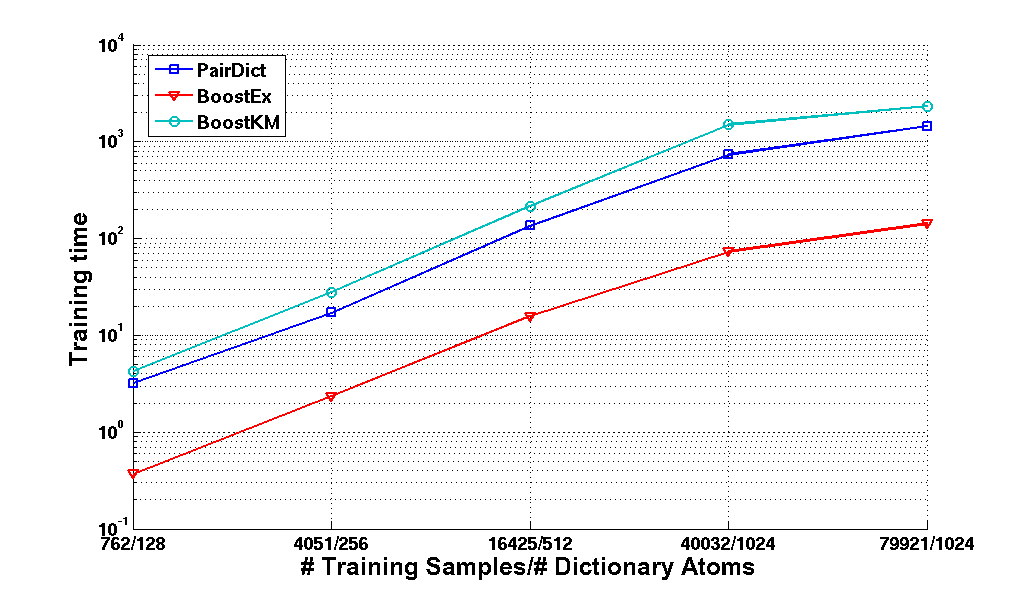}
\end{minipage}
\caption{Effect of dictionary and training set sizes on the dictionary training time for different learning schemes. The training times given are in seconds and are compared only for \textit{PairDict} ($40$ iterations), \textit{BoostEx} ($L = 50$) and \textit{BoostKM} ($L = 50$) since \textit{ExDict} and \textit{RandExAv} require no training.}
\label{Fig:SISR_dictsize}
\end{figure}

\subsection{Single Image Superresolution}
\label{sec:sr_app}
Single image superresolution (SISR) attempts to reconstruct a high-resolution image using just a single low-resolution image. It is a severely ill-posed problem and in sparse representation based approaches, the prior knowledge that natural image patches can be represented as a sparse linear combination of elementary patches, is used. The degraded test image is represented as $\mathbf{Z} =  \mathbf{\Phi}\mathbf{Y}$, where the operator $\mathbf{\Phi}$ the blurs the high-resolution image $\mathbf{Y}$ and then downsamples it. Note that $\mathbf{Y}$ and $\mathbf{Z}$ denote vectorized high- and low-resolution images respectively. Each overlapping patch obtained from the degraded image is denoted as $\mathbf{z}$. The paired dictionary learning procedure (\textit{PairDict}) proposed in \cite{yang2010image} has been very effective in recovering the high-resolution patches. This method initially creates degraded counterparts of the high-resolution training images, following which gradient-based features are extracted from the low-resolution patches and the features are appended to the corresponding vectorized high-resolution patches. These augmented features are used to train a paired dictionary $\bigl(\begin{smallmatrix}
\mathbf{D}_{lo}\\ \mathbf{D}_{hi}
\end{smallmatrix} \bigr)$
such that each low-resolution and its corresponding high-resolution patches share the same sparse code. For a low-resolution test patch $\mathbf{z}$, the sparse code $\mathbf{a}$ is obtained using $\mathbf{D}_{lo}$, and the corresponding high-resolution counterpart is recovered as $\mathbf{y} = \mathbf{D}_{hi} \mathbf{a}$. An initial estimate $\mathbf{Y}_0$ of the high-resolution image is obtained by appropriately averaging the overlapped high-resolution patches. Finally, a global reconstruction constraint is enforced by projecting $\mathbf{Y}_0$ on to the solution space of $\mathbf{\Phi}\mathbf{Y} = \mathbf{Z}$,
\begin{equation}
\min_{\mathbf{Y}} \|\mathbf{Z} - \mathbf{\Phi}\mathbf{Y}\|_2^2+ c \|\mathbf{Y} - \mathbf{Y}_0\|_2^2,
\label{eqn:SISR_backp}
\end{equation} to obtain the final reconstruction. As an alternative, the example-based procedure (\textit{ExDict}) proposed in \cite{yang2008image}, the dictionaries $\mathbf{D}_{lo}$ and $\mathbf{D}_{hi}$ are directly fixed as the features extracted from low-resolution patches and vectorized high resolution patches respectively. Similar to the \textit{PairDict} method, the global reconstruction constraint in (\ref{eqn:SISR_backp}) is imposed for the final reconstruction.

In our simulations, standard grayscale images (Table \ref{tab:SISR_compare}) are magnified by a factor of $2$, using the proposed approaches. In addition to the \textit{PairDict} and \textit{ExDict} methods, simple bicubic interpolation is also used as a baseline method. We also obtained paired dictionaries with $1024$ atoms using $100,000$ randomly chosen patches of size $5 \times 5$ from the grayscale natural images in the training set. The sparsity penalty used in training was $\lambda_{tr} = 0.15$. The training set was reduced to the size of $20,000$ samples and used as the dictionary for the \textit{ExDict method}. For ensemble learning, $L$ was fixed at $50$ and the approximation for each data sample was obtained using just a $1-$sparse representation.

For different number of training samples, we compared the training times for \textit{PairDict} ($40$ iterations), \textit{BoostEx} ($L = 50$) and \textit{BoostKM} ($L = 50$) algorithms in Figure \ref{Fig:SISR_dictsize}. The computation times reported in this paper were obtained using a single core of a 2.8 GHz Intel i7 Linux machine with 8GB RAM. The \textit{BoostKM} approach has the maximum computational complexity for training, followed by \textit{PairDict} and \textit{BoostEx} approaches. The \textit{ExDict} procedure requires no training and for \textit{RandExAv}, training time is just the time for randomly selecting $K$ samples from the training set of $T$ samples, for $L$ rounds. Clearly, the complexity incurred for this is extremely low.

\begin{figure*}[t]
\begin{minipage}[c]{0.19\linewidth}
 \centering
 \includegraphics[width = 3cm]{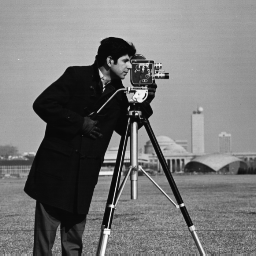}
 \centerline{{Original}}\medskip
\end{minipage}
\hfill\begin{minipage}[c]{0.19\linewidth}
 \centering
 \includegraphics[width = 1.5cm]{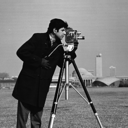}
 \centerline{{Degraded}}\medskip
\end{minipage}
\hfill\begin{minipage}[c]{0.19\linewidth}
 \centering
 \includegraphics[width = 3cm]{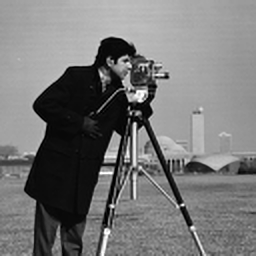}
 \centerline{{PairDict (27.78 dB)}}\medskip
\end{minipage}
\hfill\begin{minipage}[c]{0.19\linewidth}
 \centering
 \includegraphics[width = 3cm]{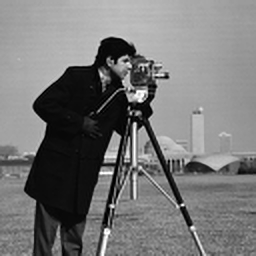}
 \centerline{{ExDict (27.78 dB)}}\medskip
\end{minipage}
\hfill\begin{minipage}[c]{0.19\linewidth}
 \centering
 \includegraphics[width = 3cm]{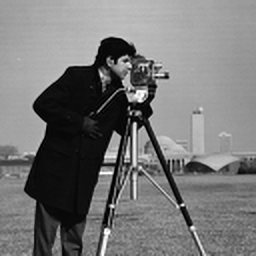}
 \centerline{{BoostKM (27.78 dB)}}\medskip
\end{minipage}
\caption{SISR of the \textit{Man} image with scaling factor of $2$. The \textit{PairDict}, \textit{ExDict}, and \textit{RandExAv} methods result in very similar high resolution images.}
% The small zoomed regions in white boxes are displayed in the lower left corner of each image.
\label{Fig:superres_recim}
\end{figure*}

For the test images, SISR is performed using the  baseline \textit{PairDict} and \textit{ExDict} approaches using a sparsity penalty of $\lambda_{te} = 0.2$. For the \textit{PairDict}, and \textit{ExDict} approaches, the code provided by the authors \cite{YangSCSR_Code}  was used to generate the results. The recovery performance of the proposed algorithms are reported in Table \ref{tab:SISR_compare}. For  \textit{PairDict}, as well the proposed ensemble methods, the dictionary size is fixed at $1024$, whereas all the examples are used for training with the \textit{ExDict} approach. We observed from our results that an ensemble representation with a simplified sparse coding scheme ($1$-sparse) matched the performance of the baseline methods (Figure \ref{Fig:superres_recim}).

\begin{table}[tb]
\renewcommand{\tabcolsep}{0.15cm}
  \centering
  \caption{Superresolution of standard images upscaled by a factor of $2$: PSNR in dB obtained with bicubic interpolation (\textit{Bicubic}), paired dictionary (\textit{PairDict}) \cite{yang2010image}, example dictionary (\textit{ExDict}) \cite{yang2008image}, BoostEx (\textit{BEx}), BoostKM (\textit{BKM}), and  RandExAv (\textit{RExAv}) methods.}
    \begin{tabular}{c|cccccc}
    \toprule
    \textbf{Image} & \textbf{Bicubic} & \textbf{PairDict} & \textbf{ExDict} & \textbf{BEx} & \textbf{BKM} & \textbf{RExAv} \\
    \midrule
    Lena  & 34.10 & \textbf{35.99} & \textbf{35.99} & 35.97 & 35.95 & \textbf{35.99} \\
    Boat  & 29.94 & \textbf{31.34} & \textbf{31.34} & 31.28 & 31.23 & 31.29 \\
    House & 32.77 & \textbf{34.49} & \textbf{34.49} & 34.38 & 34.41 & 34.41 \\
    Cameraman & 26.33 & \textbf{27.78} & \textbf{27.78} & 27.72 & \textbf{27.78} & 27.71 \\
    Straw & 24.20 & 25.93 & 25.93 & 25.90 & 25.90 & \textbf{25.94} \\
    Girl  & 33.81 & \textbf{35.39} & \textbf{35.39} & 35.33 & 35.37 & 35.35 \\
    \bottomrule
    \end{tabular}%
  \label{tab:SISR_compare}
\end{table}%

\section{Application: Unsupervised Clustering}
\label{sec:cluster_app}
Conventional clustering algorithms such as K-Means provide good clusterings only when the natural clusters of the data are distributed around a mean vector in space. For data that lie in a union of low-dimensional subspaces, it is beneficial to develop clustering algorithms that try to model the actual data distribution better. The sparse subspace clustering method \cite{elhamifar2012sparse}, a special case of which is referred to as the $\ell_1$ graph clustering  \cite{Cheng2010}, results in clusters that correspond to subspaces of data. This is achieved by representing each example as a sparse linear combination of the others and finally performing spectral clustering using a similarity matrix obtained from the coefficient matrix. The clustering method has the advantage of incorporating the noise model directly when performing sparse coding, thereby achieving robustness. The coefficient vector for the $i^{\text{th}}$ data sample is obtained as
\begin{equation}
\min_{\mathbf{b}_i} \left\|\mathbf{x}_i-\mathbf{X} \mathbf{a}_i\right\|+\lambda \|\mathbf{a}_i\|_1, \text{ subj. to. } a_{ii} = 0.
\label{eqn:l1graph_coeff}
\end{equation} By imposing the constraint that the $i^{\text{th}}$ element of $\mathbf{a}_i$ should be $0$, we ensure that a data sample is not represented by itself, which would have resulted in a trivial approximation. The coefficient matrix is denoted as $\mathbf{A} = [\mathbf{a}_1 \mathbf{a}_2 \ldots \mathbf{a}_T]$, and  spectral clustering \cite{ng2002spectral} is performed by setting the similarity matrix to the symmetric non-negative version of the coefficient matrix, $\mathbf{S} = |\mathbf{A}|+|\mathbf{A}^T|$. Computing the graph in this case necessitates the computation of sparse codes of $T$ data samples with a $M \times (T-1)$ dictionary. Sparse coding-based graphs can also constructed based on coefficients obtained with a dictionary $\mathbf{D}$, inferred using the \textit{Alt-Opt} procedure. Denoting the sparse codes for the examples $\mathbf{X}$ by the coefficient matrix $\mathbf{A} = [\mathbf{a}_1 \ldots \mathbf{a}_T]$, the similarity matrix can be constructed as $\mathbf{S} = |\mathbf{A}^T \mathbf{A}|$. Similar to the $\ell_1$ graphs, this similarity matrix can be used with spectral clustering to estimate the cluster memberships \cite{Ramirez}. In this case, the dominant complexity in computing the graph is in learning the dictionary, and obtaining the sparse codes for each example. When the number of training examples is large, or when the data is high-dimensional, approaches that use sparse coding-based graphs incur high computational complexity. 

We propose to construct sparse representation-based graphs using our ensemble approaches and employ them in spectral clustering. In our ensemble approaches, we have two example-based procedures, (\textit{RandExAv} and \textit{BoostEx}) and one that uses K-Means dictionaries (\textit{BoostKM}). For \textit{BoostKM}, we obtain $L$ dictionaries of size $K$ using the boosting procedure, with $1-$sparse approximations. The final coefficient vector of length $LK$ for the data sample $\mathbf{x}_i$ is obtained as, $\mathbf{a}_i = [\mathbf{a}_{1,i}^T \mathbf{a}_{2,i}^T \ldots \mathbf{a}_{L,i}^T]^T$, where $\mathbf{a}_{l,i}$ is the coefficient vector for round $l$. The similarity matrix is then estimated as  $\mathbf{S} = |\mathbf{A}^T \mathbf{A}|$. In the example-based procedures, again $1-$sparse representation is used to obtain the coefficient vectors $\{\mathbf{a}_{1,i},\mathbf{a}_{2,i} \ldots, \mathbf{a}_{L,i}\}$, for a data sample $\mathbf{x}_i$. A cumulative coefficient vector of length $T$ can be obtained by recognizing that each coefficient in $\mathbf{a}_{l,i} \in \mathbb{R}^K$, can be associated to a particular example, since $\mathbf{D}_l$ is an example-based dictionary. Therefore a new $1-$sparse coefficient vector $\bar{\mathbf{a}}_{l,i} \in \mathbb{R}^T$ is created such that $\mathbf{D}_l\mathbf{a}_{l,i} = \bar{\mathbf{X}}\bar{\mathbf{a}}_{l,i}$, where $\bar{\mathbf{X}}$ contains the normalized set of data samples $\mathbf{X}$. Finally the cumulative coefficient vector for $\mathbf{x}_i$ is obtained as $\sum_{l=1}^L \beta_l \bar{\mathbf{a}}_{l,i}$. They are then stacked to form the coefficient matrix $\bar{\mathbf{A}} = [\bar{\mathbf{a}}_1 \ldots \bar{\mathbf{a}}_T]$. Spectral clustering can be now performed using the similarity matrix $\mathbf{S} = |\bar{\mathbf{A}}| + |\bar{\mathbf{A}}^T|$. The clustering performance was evaluated in terms of accuracy and normalized mutual information (NMI), and compared with $\ell_1$ graphs. As seen from Table \ref{table:clusperf}, the ensemble-based approaches result in high accuracy as well as NMI. In all our simulations, data was preprocessed by centering and normalizing to unit norm. It was observed that the proposed ensemble methods incur comparable computational complexity to $\ell_1$ graphs for datasets with small data dimensions. However, we observed significant complexity reduction with the \textit{USPS} dataset, which contains $9298$ samples of $256$ dimensions. To cluster the \textit{USPS} samples, the $\ell_1$ graph approach took $411.85$ seconds to compute the sparse codes, whereas \textit{BoostEx}, \textit{RandExAv}, and \textit{BoostKM} took $152.56$, $147.93$, and $83.58$ seconds respectively. This indicates the suitability of the proposed methods for high-dimensional, large scale data. 

\begin{table}[tb]
\renewcommand{\tabcolsep}{0.15cm}
  \centering
  \caption{Comparison of the clustering performances (accuracy and normalized mutual information) of the algorithms with standard datasets. The best maximum or average performance is given in bold font.}
    \begin{tabular}{c|cc|cc|cc|cc}
    \toprule
    \multicolumn{1}{l}{\multirow{2}[0]{*}{\textbf{Dataset}}} & \multicolumn{2}{|c|}{\textbf{$\ell_1$ graph}} & \multicolumn{2}{|c|}{\textbf{BoostEx}} & \multicolumn{2}{|c|}{\textbf{RandExAv}} & \multicolumn{2}{|c}{\textbf{BoostKM}} \\
    \multicolumn{1}{l}{} & \multicolumn{1}{|c}{max} & \multicolumn{1}{c|}{avg} & \multicolumn{1}{|c}{max} & \multicolumn{1}{c|}{avg} & \multicolumn{1}{|c}{max} & \multicolumn{1}{c|}{avg} & \multicolumn{1}{|c}{max} & \multicolumn{1}{c}{avg} \\
    \midrule
    \multicolumn{1}{c}{}& \multicolumn{2}{c}{} & \multicolumn{2}{c}{\textbf{Accuracy}} &\multicolumn{2}{c}{} & \multicolumn{2}{c}{} \\
    \midrule
    Digits & \textbf{88.72} & 74.61 & 88.63 & 76.79 & 88.62 & \textbf{77.56} & 88.31 & 76.85 \\
    Soybean & 67.44 & 58.22 & 67.26 & 63.21 & \textbf{70.82} & \textbf{65.31} & 69.22 & 63.50 \\
    Segment & 65.32 & 57.84 & 63.42 & 58.46 & 63.33 & 56.49 & \textbf{65.63} & \textbf{58.67} \\
    Satimage & 77.56 & 69.37 & 75.03 & 72.88 & \textbf{83.59} & \textbf{75.25} & 71.62 & 65.32 \\
    USPS & 78.24 & 62.47 & 75.01 & 68.18 & 78.26 & 64.14 & \textbf{90.96} & \textbf{75.18} \\
    \midrule
        \multicolumn{1}{c}{}& \multicolumn{2}{c}{} &\multicolumn{2}{c}{\textbf{NMI}} & \multicolumn{2}{c}{} & \multicolumn{2}{c}{} \\
    \midrule
        Digits & \textbf{84.97} & 76.50 & 84.69 & 77.67 & 84.73 & \textbf{78.04} & 84.31 & 78.02 \\
    Soybean & 74.55 & 65.94 & 72.84 & 68.66 & \textbf{77.50} & \textbf{73.71} & 76.05 & 71.66 \\
    Segment & \textbf{59.14} & 53.91 & 56.72 & 53.23 & 54.94 & 51.20 & 58.44 & \textbf{55.32} \\
    Satimage & 65.09 & 61.20 & 62.98 & 59.01 & \textbf{69.38} & \textbf{66.12} & 60.27 & 55.57 \\
    USPS  & 81.04 & 74.14 & 70.52 & 66.89 & 82.67 & 76.95 & \textbf{82.78} & \textbf{78.37} \\
    \bottomrule
    \end{tabular}%
  \label{table:clusperf}%
\end{table}%

\section{Conclusions}
\label{sec:conclusions}
We proposed and analyzed the framework of ensemble sparse models, where the data is represented using a linear combination of approximations from multiple sparse representations. Theoretical results and experimental demonstrations show that an ensemble representation leads to a better approximation when compared to its individual constituents. Three different methods for learning the ensemble were proposed. Results in compressive recovery showed that the proposed approaches performed better than the baseline sparse coding method. Furthermore, the ensemble approach performed comparably to several recent techniques in single image superresolution. Results with unsupervised clustering also showed that the proposed method leads to better clustering performance in comparison to the $\ell_1$ graph method.

\bibliographystyle{IEEEtran}
\bibliography{strings}
\end{document}